\documentclass[runningheads]{llncs}
\usepackage[T1]{fontenc}
\usepackage{graphicx,verbatim}
\usepackage{amsmath}
\usepackage{amsfonts}
\usepackage{booktabs}
\usepackage{multirow}
\usepackage{colortbl}
\usepackage{xcolor}
\usepackage{hyperref}
\usepackage{tabularx}

\begin{document}
\title{DLA-Count: Dynamic Label Assignment Network for Dense Cell Distribution Counting}
\titlerunning{DLA-Count: Dynamic Label Assignment Network}
\begin{comment}  %% Removed for anonymized MICCAI 2025 submission
\author{First Author\inst{1}\orcidID{0000-1111-2222-3333} \and
Second Author\inst{2,3}\orcidID{1111-2222-3333-4444} \and
Third Author\inst{3}\orcidID{2222--3333-4444-5555}}
%
\authorrunning{F. Author et al.}
\institute{Princeton University, Princeton NJ 08544, USA \and
Springer Heidelberg, Tiergartenstr. 17, 69121 Heidelberg, Germany
\email{lncs@springer.com}\\
\url{http://www.springer.com/gp/computer-science/lncs} \and
ABC Institute, Rupert-Karls-University Heidelberg, Heidelberg, Germany\\
\email{\{abc,lncs\}@uni-heidelberg.de}}

\end{comment}

\author{Yuqing Yan, Yirui Wu*}  %% Added for anonymized MICCAI 2025 submission
%\authorrunning{Anonymized Author et al.}
\institute{Hohai University}

\maketitle              % typeset the header of the contribution
\begin{abstract}
% 第一处 应该提升模型的鲁棒性
% 第二处 Dynamic Neighborhood Gaussian Matching (DNGM) 是K紧邻动态匹配 可以叫做K紧邻匈牙利匹配算法，这个算法的作用是鲁棒性！ 如果使用了高斯函数作为权重的话 可以！
% 摘要尽可能简单，你只说提出了XX，作用是XXX。然后在后面再详细介绍
% 文章要这样写： 首先介绍这个工作，然后介绍前人的工作，他们的缺点是什么 再说自己为了弥补这个缺点做了哪些工作
%提升了多少 最好体现出来！ 有没有拿到SOTA
% Cell counting in microscopic images, as a critical technology in medical image analysis and quantitative biology research, directly impact the reliability and reproducibility of various studies.
Cell counting remains a fundamental yet challenging task in medical and biological research due to the diverse morphology of cells, their dense distribution, and variations in image quality.
% Challenges such as densely distributed cells, significant morphological heterogeneity, and complex scale variations remain persistent issues in cell counting tasks.
We present DLA-Count, a breakthrough approach to cell counting that introduces three key innovations: (1) K-adjacent Hungarian Matching (KHM), which dramatically improves cell matching in dense regions, (2) Multi-scale Deformable Gaussian Convolution (MDGC), which adapts to varying cell morphologies, and (3) Gaussian-enhanced Feature Decoder (GFD) for efficient multi-scale feature fusion. Our extensive experiments on four challenging cell counting datasets (ADI, MBM, VGG, and DCC) demonstrate that our method outperforms previous methods across diverse datasets, with improvements in Mean Absolute Error of up to 46.7\% on ADI and 42.5\% on MBM datasets. Our code is available at \href{https://anonymous.4open.science/r/DLA-Count}{https://anonymous.4open.science/r/DLA-Count}.
% To address these challenges, we introduce the Dynamic Label Assignment Network (DLA-Count) that significantly enhances the robustness of the model and improves counting accuracy through three key innovations: First, the K-adjacent Hungarian Matching (KHM) Algorithm is proposed, which employs k-nearest neighbor-derived Gaussian attenuation to construct adaptive cost matrices, significantly enhancing matching robustness against dense distributions and spatial ambiguities. Moreover, we designed the Multi-scale Deformable Gaussian Convolution (MDGC), which integrates depthwise separable convolution with parameterized Gaussian kernel prediction, dynamically fusing multi-scale features via channel attention to enhance structural adaptability against morphological heterogeneity and scale variations.
% Building on this, the Gaussian-enhanced Feature Decoder (GFD) optimizes the feature pyramid architecture through a gated fusion strategy, enhancing multi-level geometric information preservation. Experimental results on multiple benchmark datasets demonstrate that DLA-Count achieves state-of-the-art performance in terms of MAE and MSE, outperforming existing approaches on DCC, ADI, MBM, and VGG. Our code is available in \href{https://anonymous.4open.science/r/DLA-Count}{https://anonymous.4open.science/r/DLA-Count}.
%提升了多少 最好体现出来！ 有没有拿到SOTA
% Authors must provide keywords and are not allowed to remove this Keyword section.

\end{abstract}
\section{Introduction}
% 在哪些地方应用要加引用
% 很多地方有一些格式问题，对照我们的baseline完善一下
% SOTA应该说最近的方法

% 文章要这样写： 首先介绍这个工作，然后介绍前人的工作，他们的缺点是什么 再说自己为了弥补这个缺点做了哪些工作
Accurate cell counting is a fundamental task in medical and biological research, with critical value for cancer diagnosis \cite{dia}, drug screening \cite{covid19}, and regenerative medicine \cite{blood}. In tumor pathology, precise assessment of cell density and distribution patterns directly impacts disease grading and treatment decisions. in drug development, accurate monitoring of cell proliferation rates serves as a key indicator for evaluating drug efficacy. Traditional manual counting methods are not only time-consuming and laborious but also susceptible to human subjective factors, leading to issues with consistency and reproducibility. Automated cell counting techniques aim to address these problems but still face three major challenges: first, the high morphological variability of cells, with significant differences in size, shape, and density across different tissues and pathological states \cite{mic}; second, cell overlap and boundary ambiguity in dense regions, making precise identification difficult \cite{auto}; finally, variations in imaging conditions (such as brightness, contrast, and noise levels) further increase the complexity of analysis.

Early automated methods like FCN-based regression (e.g., Count-ception \cite{adi}) and detection-driven frameworks (e.g., Faster R-CNN \cite{frcnn}) laid foundational solutions but struggled with dense cell clusters and morphological diversity. Recent advancements in cell-specific models—including point detection networks (e.g., Cellpose \cite{cellpose})—have improved performance but still face critical limitations. Global matching algorithms in popular tools (e.g., StarDist \cite{stardist}) frequently misidentify overlapping cells due to fixed-distance thresholds, while standard convolutions in architectures like U-Net fail to adapt to irregular cell shapes. Even state-of-the-art approaches \cite{pgc}\cite{dcl} exhibit degraded precision in extreme-density scenarios, primarily due to information loss during multi-scale feature fusion.

\begin{figure}[!t]
    \centering
    \includegraphics[width=\textwidth]{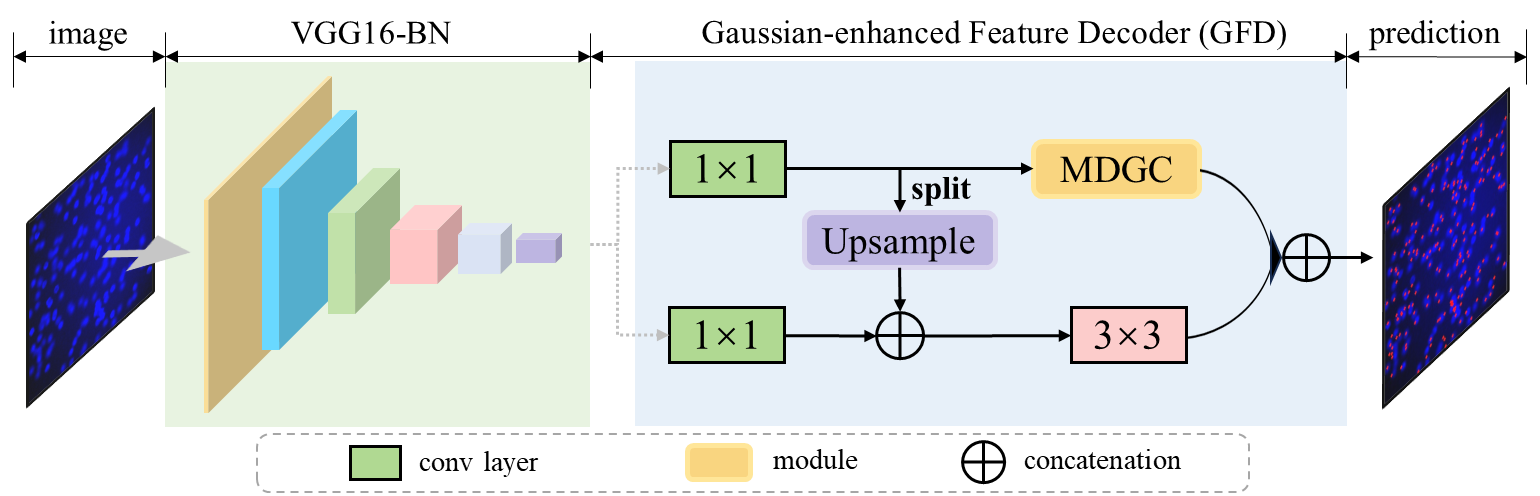}
    \caption{Overall architecture diagram of DLA-Count.}
    \label{model}
\end{figure}

To overcome these limitations, we propose the Dynamic Label Assignment Network (DLA-Count) which systematically resolves key challenges above in dense cell analysis, shown in Fig.\ref{model}. This model design adopts an encoder-decoder architecture, first uses a pre-trained VGG16 -
bn \cite{vgg16bn} as the encoder to extract multi-scale features from
the input image, which are then fed into the proposed Gaussian-enhanced Feature Decoder (GFD). Our contributions can be summarized as follows:
\begin{enumerate}
    \item We propose the K-adjacent Hungarian Matching (KHM) Algorithm, which enables context-aware label assignment through adaptive radius calculation and geometric-semantic cost optimization.
    \item The Multi-scale Deformable Gaussian Convolution (MDGC) dynamically adjusts kernel shapes and scales to capture diverse cellular morphologies. On this basis, we introduce the Gaussian-enhanced Feature Decoder (GFD), which preserves geometric fidelity via selective multi-scale fusion and spatial attention mechanisms.
    \item Extensive experiments demonstrate state-of-the-art performance across multiple benchmark datasets, outperforming existing methods in both counting accuracy and robustness precision under complex biological scenarios.
\end{enumerate}

\begin{figure}[!t]
\includegraphics[width=\textwidth]{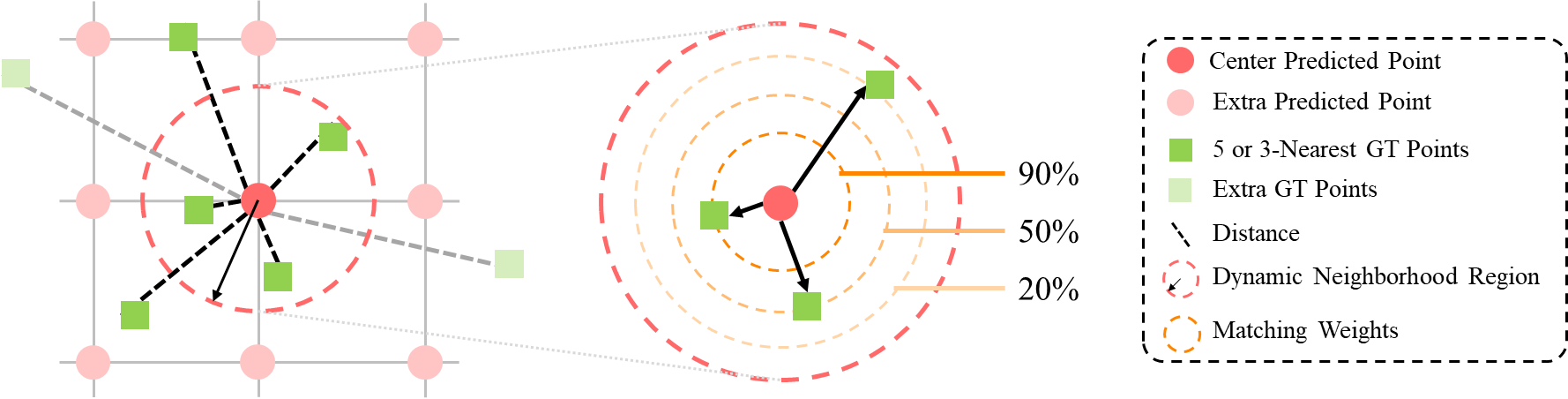}
\caption{Core design of the K-adjacent Hungarian Matching (KHM) Algorithm.} \label{matching}
\end{figure}
\section{Method}
% 首先介绍整体架构！ 输入是什么，输出是什么？  如何计算损失？ 
% 然后各个创新点介绍！
\subsection{Problem Definition and Method Overview}

\subsubsection{Problem Definition}
The cell counting and localization problem can be formally defined as follows: given an input image $I \in \mathbb{R}^{H \times W \times 3}$, where $H$ and $W$ represent the height and width of the image, our goal is to predict the set of cell positions $P = \{p_1, p_2, ..., p_N\}$, where $N$ is the total number of cells and $p_i = (x_i, y_i)$ represents the center coordinates of the $i$-th cell. Unlike general object detection, the cell counting problem presents three unique challenges: (1) cells are typically small in size and numerous (a single image may contain hundreds or even thousands of cells); (2) cell morphology exhibits significant variability, including differences in size, shape, and density; (3) cells may overlap in dense regions, with ambiguous boundaries. These characteristics make the task of precisely detecting the location of each cell challenging.

In a standard supervised learning setting, we have a training dataset $\mathcal{D} = \{(I_i, G_i)\}_{i=1}^M$, where $G_i = \{g_1, g_2, ..., g_{N_i}\}$ represents the ground truth locations of $N_i$ cells in image $I_i$. Our goal is to learn a mapping function $f_\theta: I \mapsto P$ that minimizes the difference between the predicted positions $P$ and the ground truth positions $G$.

\subsubsection{Method Overview}
As shown in Figure~\ref{model}, our proposed DLA-Count framework adopts an end-to-end design without the need for complex post-processing steps. Given an input image $I \in \mathbb{R}^{H \times W \times 3}$, we first use VGG16-bn as the backbone network to extract multi-scale features. Specifically, we extract feature maps from the fourth and fifth stages of the network, denoted as $\{F_4, F_5\}$, with corresponding channel dimensions of $\{512, 512\}$ and spatial resolutions of $\{1/8, 1/16\}$ of the input image. These two levels of features capture medium-scale structural features and large-scale semantic information of cells, respectively.

The extracted features are subsequently fed into the Gaussian-enhanced Feature Decoder (GFD) for processing. The GFD module adaptively fuses multi-scale features to generate a high-resolution feature map for cell positions, $F_{out} \in \mathbb{R}^{H/8 \times W/8 \times C}$, where $C$ is the number of feature channels. GFD employs a Gaussian enhancement strategy to effectively strengthen cell region features while suppressing background interference.

Finally, we apply a simple regression head (consisting of a 1×1 convolutional layer) to transform the fused feature map into a cell position prediction map $P_{map} \in \mathbb{R}^{H/8 \times W/8 \times 1}$. Local maxima in the position prediction map are extracted as the predicted cell center coordinate set $P = \{p_1, p_2, ..., p_N\}$, where $N$ is the total number of detected cells.

To compute the training loss, we need to match the predicted point set $P$ with the ground truth point set $G$. This is essentially a bipartite graph matching problem, traditionally implemented using the standard Hungarian algorithm. However, the standard Hungarian algorithm uses a fixed-threshold global matching strategy, which is difficult to adapt to images with significant variations in cell density. Therefore, we propose the KHM algorithm, which dynamically adjusts matching thresholds by considering local cell density, significantly improving matching accuracy, especially in regions with non-uniform cell density.

In the following sections, we will provide detailed descriptions of the two core innovative modules of DLA-Count: the KHM and MDGC. These modules work collaboratively to address the challenges of morphological diversity and density variations in cell counting tasks.

\subsection{K-adjacent Hungarian Matching (KHM) Algorithm}
The KHM algorithm addresses fundamental limitations of traditional matching methods in dense cell counting scenarios through a spatially-aware Hungarian matching approach, as illustrated in Fig.~\ref{matching}. Unlike conventional fixed-radius or global-density strategies, KHM implements an adaptive matching paradigm that responds dynamically to local density variations.

For each predicted point $p_i$, KHM computes a density-adaptive search radius:
\begin{equation}
\delta_i = \frac{1}{k} \sum_{m=1}^k D_{i,m}^{(k)} \quad
\end{equation}
where $D_{i,m}^{(k)}$ represents the distance to the $k$-th nearest ground truth point. This mathematically principled approach automatically contracts $\delta_i$ in densely populated regions to prevent false matches while expanding it in sparse areas.

Within each adaptive radius, KHM employs a geometry-sensitive cost function with Gaussian decay:
\begin{equation}
w_{i,j} = \exp\left(-\frac{D_{i,j}^2}{2\sigma^2}\right) \quad 
\end{equation}
%This formulation prioritizes matches proximal to neighborhood centers while preserving unmodified distance relationships outside the $\delta_i$ boundary.
where \(w_{i,j}\) represents the matching weight between prediction \(p_i\) and ground truth \(g_j\). This weight decays exponentially with squared distance \(D_{i,j}^2\), implementing soft spatial constraints within the adaptive radius $\delta_i$. The weights form an assignment matrix where each element quantifies match quality. During Hungarian matching, higher weights indicate geometrically favorable pairings, effectively prioritizing matches near neighborhood centers while preserving original distance relationships beyond $\delta_i$.

%The algorithm synthesizes a composite cost matrix:
%\begin{equation}
%C = 1.2 C^{\text{point}} + 0.8 C^{\text{class}}
%\end{equation}
%which optimally balances spatial precision with semantic consistency. This dual-metric approach overcomes the inherent limitations of single-criterion matching systems, particularly in challenging scenarios with overlapping cells where traditional methods typically underperform.

Through these systematic, formula-driven adaptations, KHM achieves density-agnostic matching without requiring extensive hyperparameter tuning, demonstrating superior performance compared to both conventional global matching algorithms and heuristic adaptive techniques.

\subsection{Multi-scale Deformable Gaussian Convolution (MDGC)}
The MDGC module redefines feature extraction for cellular imaging by combining deformable convolution principles with morphology-specific Gaussian adaptation, addressing three key limitations of existing methods. Unlike standard deformable convolutions that learn coordinate offsets alone, MDGC introduces parametric anisotropic Gaussian kernels dynamically generated through a light-weight subnetwork. This network predicts three parameters per spatial location: base standard deviation \(\sigma \in [1, 10]\), and center offsets \(\Delta x,\Delta y \in [-2, 2]\). During training, \(\sigma, \Delta x, \Delta y\) along with $s_x$ and $s_y$ are jointly optimized through backpropagation, where the lightweight subnetwork learns to predict these parameters in an end-to-end manner guided by gradient descent. Crucially, the elliptical kernel shape is governed by \(\sigma_x = s_x \cdot \sigma\) and \(\sigma_y = s_y \cdot \sigma\), where $s_x$ and $s_y$ are trainable axial scaling factors that convert the base $\sigma$ into anisotropic deviations, formally expressed as:
\begin{equation}
    \mathcal{K}(u,v) = \frac{1}{2\pi\sigma_x\sigma_y} \exp\left( -\frac{(u-\Delta x)^2}{2\sigma_x^2} - \frac{(v-\Delta y)^2}{2\sigma_y^2} \right)
\end{equation}
This allows the kernel to adapt to elliptical cell shapes while maintaining stable gradient propagation, unlike rigid offset-based deformable convolutions.

\begin{figure}[!t]
\includegraphics[width=\textwidth]{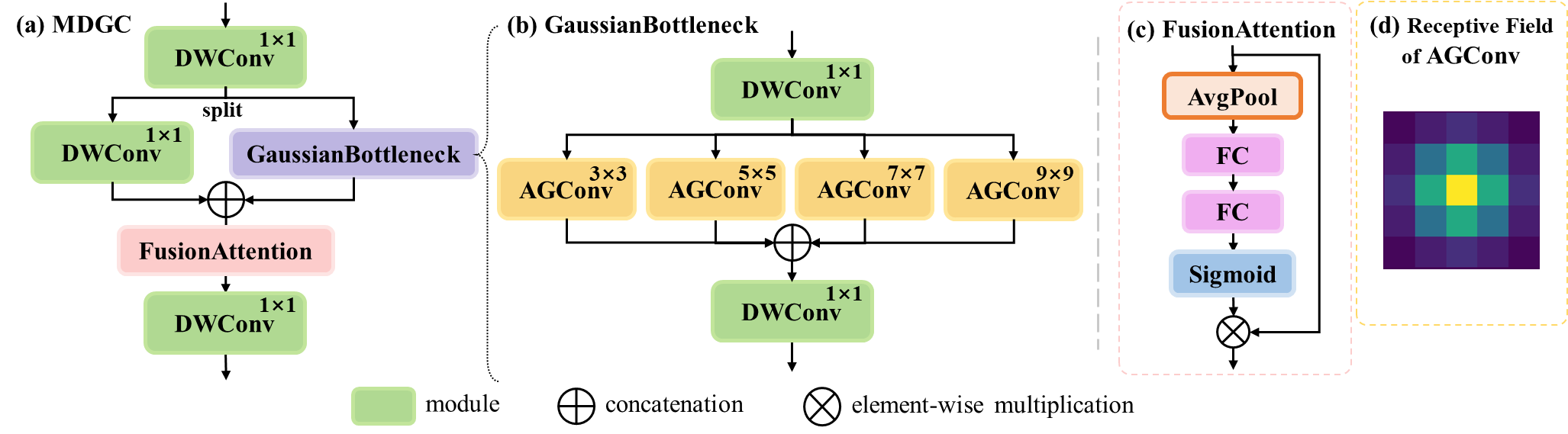}
\caption{The architecture of our Multi-scale Deformable Gaussian Convolution (MDGC)-(a), internal modules-(b)(c)(d). (a) shows the overall MDGC structure with parallel processing paths; (b) illustrates the Gaussian Bottleneck with multi-scale AGConv operations; (c) details the FusionAttention mechanism for adaptive feature recalibration; and (d) visualizes the receptive field pattern generated by the Adaptive Gaussian Convolution (AGConv).} \label{gauconv}
\end{figure}

The architecture, shown in Fig.\ref{gauconv}, employs dual-path multi-scale processing:
A standard path with depthwise separable convolutions captures regular structures; 
A Gaussian path applies the deformable kernels at matching scales with \(\sigma_x, \sigma_y\). 
Feature fusion uses an enhanced channel attention mechanism on concatenated multi-scale features \(F_{\text{cat}}\). By incorporating residual connections into weight generation:
\begin{equation}
    \alpha' = \alpha + \text{GAP}(F_{\text{cat}})
\end{equation}
the module preserves original feature semantics while prioritizing discriminative patterns, outperforming conventional deformable convolutions in capturing irregular cell morphologies with fewer artifacts.

%The MDGC module represents a paradigm shift in convolutional feature extraction for cellular image analysis, addressing three fundamental challenges through its innovative architecture. The first component—deformable Gaussian kernel generation—employs a lightweight parametric subnetwork that predicts crucial kernel characteristics in real-time. This subnetwork, comprising global average pooling followed by two 1$\times$1 convolutional layers with SiLU activation, outputs a triplet \((\sigma,\Delta x,\Delta y)\) for each spatial position, where $\sigma$ controls the kernel's standard deviation (constrained to [1,10]) and \(\Delta x,\Delta y\) represent learnable center offsets bounded within [-2,2] pixels. These parameters enable the dynamic generation of anisotropic Gaussian kernels formalized as:
% \begin{equation}
%    \mathcal{K}(u,v) = \frac{1}{2\pi\sigma_x\sigma_y} \exp\left( -\frac{(u-\Delta x)^2}{2\sigma_x^2} - \frac{(v-\Delta y)^2}{2\sigma_y^2} \right)
% \end{equation}
%where \(\sigma_x\) and \(\sigma_y\) are derived from the predicted $\sigma$ through axial scaling factors learned during training, allowing the kernel to adapt to elliptical cell morphologies.

\begin{table}[!t]
\centering
\caption{Overview of four public datasets and their features.}\label{datasets}
\begin{tabular}{lccc}
\hline
Datasets & Image Size & Cell Count & Type \\
\hline
ADI  & 150$\times$150 & 165 $\pm$ 44 & real \\
\hline
MBM  & 600$\times$600 & 126 $\pm$ 33 & real \\
\hline
VGG  & 256$\times$256 & 174 $\pm$ 64 & synthetic \\
\hline
DCC  & 306$\times$322 $\sim$ 798$\times$788 & 34 $\pm$ 22 & real \\
\hline
\end{tabular}
\end{table}

\section{Experiments}
\subsection{Experimental Setups}
\subsubsection{Datasets.}
We validate our methodology on four publicly available biomedical imaging benchmarks shown in Tab.\ref{datasets}:
    %\item Human Subcutaneous Adipose Tissue (ADI) dataset \cite{adi}: Developed from high-resolution tissue scans using a 1700 $\times$ 1700 sampling window, ADI contains 150 $\times$ 150 image regions showing tightly packed fat cells with diverse sizes. The cells occupy over 90\% of the image area, creating challenging counting conditions due to minimal gaps between adjacent cells.
    
ADI dataset \cite{adi}: High-resolution subcutaneous adipose images (150 $\times$ 150) with densely packed adipocytes (90\% coverage), minimal intercellular gaps;
    %\item Modified Bone Marrow (MBM) dataset \cite{mbm}: Consisting of 11 large 1200 $\times$ 1200 microscope images (analyzed as 600 $\times$ 600 sections) from 8 donors, MBM uses blue-pink staining to clearly distinguish cell nuclei from surrounding material. It provides insights into different blood cell types within bone marrow tissue.    
MBM dataset \cite{mbm}: Bone marrow microscopy (original 1200 $\times$ 1200, analyzed as 600 $\times$ 600 sections), blue-pink stained nuclei from 8 donors;
    %\item Synthetic Fluorescence Microscopy (VGG) dataset \cite{vgg}: This computer-generated dataset mimics fluorescence microscope images (256 $\times$ 256) of bacterial colonies. Each image contains 174 $\pm$ 64 overlapping cells distributed across different depth levels, simulating real-world microscopy challenges.    
VGG dataset \cite{vgg}: Synthetic fluorescence microscopy (256 $\times$ 256), simulates 174 $\pm$ 64 bacteria colonies with multi-layer overlaps;
    %\item Dublin Cell Counting (DCC) dataset \cite{dcc}: Featuring three cell types - mouse stem cells, human lung cancer cells, and human immune cells - DCC includes varied image sizes (306 $\times$ 322 to 798 $\times$ 788 pixels) to test counting methods across different image sizes and cell characteristics.
    %To address limited data availability, we divided each dataset into training and test sets using a 4 : 1 ratio.    
DCC dataset \cite{dcc}: Multi-cell dataset (stem/ cancer/ immune cells) with varied image sizes (306 $\times$ 322 to 798 $\times$ 788) for scale robustness testing.
\subsubsection{Implementation Details.}
Given the varying image sizes across the four datasets and the need for accelerated training, we implemented the following unified preprocessing pipeline:
Converted original grayscale PNG annotations into lightweight TXT coordinate files. We extracted \((x,y)\) coordinates of all non-background color pixels from each annotation image, storing them as integer values in row-major order. 
%This transformation achieves three optimization objectives:
%Eliminates impact from original image size variations; Reduces storage footprint by $\sim$80\%; Accelerates annotation loading during training. %The generated TXT files maintain strict correspondence with original images, where each coordinate represents the center location of a target cell.
%This multi-scale geometric transformation strategy amplifies data diversity while retaining diagnostically critical morphological features for biomedical imaging analysis.

The experiments were conducted on a computing platform equipped with an NVIDIA RTX 4090 GPU with 24GB of memory, and were implemented in PyTorch. For optimization, we employed the Adam optimizer with an initial learning rate of 1e-4 and a batch size of 4.
To enhance model's generalization ability and robustness, we implement sequential data augmentation including random flipping and random cropping.

\begin{table}[!t]
\centering
\caption{Comparison with state-of-the-art approaches on four public
datasets. Performance is measured by MAE and MSE.}\label{counting}
\begin{tabular}{lcccccccc}
\toprule
\multirow{2}{*}{Approaches} & \multicolumn{2}{c}{ADI} & \multicolumn{2}{c}{MBM} & \multicolumn{2}{c}{VGG} & \multicolumn{2}{c}{DCC} \\
\cmidrule(lr){2-3} \cmidrule(lr){4-5} \cmidrule(lr){6-7} \cmidrule(lr){8-9}
& MAE & MSE & MAE & MSE & MAE & MSE & MAE & MSE \\
\midrule
MCNN \cite{mcnn} & 25.8 & 35.7 & 3.2 & 4.3 & 20.9 & 25.3 & 5.4 & 6.4 \\
FCRN \cite{fcrn} & 20.6 & 28.3 & 2.8 & 3.7 & 17.7 & 21.5 & 5.6 & 7.3 \\
CSRNet \cite{csrnet} & 13.5 & 18.3 & 2.2 & 2.9 & 7.9 & 10.2 & 2.2 & 2.9 \\
SFCN \cite{sfcn} & 16.0 & 22.3 & 2.4 & 3.1 & 13.8 & 17.9 & 2.7 & 3.7 \\
DMCount \cite{dmcount} & 9.4 & 13.5 & 2.6 & 3.5 & 6.0 & 8.0 & 2.6 & 3.8 \\
SASNet \cite{sasnet} & 9.0 & 12.2 & 3.9 & 5.2 & 4.9 & 6.8 & 8.9 & 12.0 \\
DQN \cite{dqn} & 9.7 & 13.1 & 3.1 & 4.2 & 5.5 & 7.3 & 3.5 & 4.6 \\
OrdinalEntropy \cite{ordientro} & 9.1 & 12.0 & 2.9 & 3.8 & 5.7 & 7.8 & 3.2 & 4.3 \\
DiffuseDenoiseCount \cite{ddc} & 8.8 & 11.9 & 2.9 & 3.9 & 5.5 & 7.0 & 2.8 & 3.7 \\
DCL \cite{dcl} & 8.4 & 11.7 & 1.4 & 2.1 & 4.1 & 5.9 & 0.8 & 1.3 \\
\textbf{Ours} & \textbf{4.1} & \textbf{5.7} & \textbf{0.1} & \textbf{0.3} & \textbf{2.0} & \textbf{2.7} & \textbf{1.7} & \textbf{3.0} \\
\midrule
\rowcolor{gray!20}
Improvements  & 51.2\% & 51.3\% & 92.9\% & 85.7\% & 51.2\% & 54.2\% & - & - \\
\bottomrule
\end{tabular}
\end{table}
\begin{minipage}{\textwidth}
\begin{minipage}[t]{0.3\textwidth}
\makeatletter\def\@captype{table}
\caption{Model ablation experiment on ADI.}\label{ablation}
\begin{tabular}{lcc}
\toprule
Module & MAE & MSE \\
\midrule
Baseline & 9.0 & 13.1 \\
+KHM & 4.6 & 6.1 \\
+MDGC & 4.4 & 5.9 \\
\midrule
\textbf{+All} & \textbf{4.1} & \textbf{5.7} \\
\bottomrule
\end{tabular}
\end{minipage}
\begin{minipage}[t]{0.33\textwidth}
\makeatletter\def\@captype{table}
\caption{KHM ablation experiment on ADI. }\label{kablation}
\begin{tabular}{lcc}
\toprule
Parameter & MAE & MSE \\
\midrule
k=3 & 4.7 & 6.2 \\
k=5 & 4.6 & 6.1 \\
\midrule
\textbf{Full Model} & \textbf{4.1} & \textbf{5.7} \\
\bottomrule
\end{tabular}
\end{minipage}
\begin{minipage}[t]{0.31\textwidth}
\makeatletter\def\@captype{table}
\caption{MDGC ablation experiment on ADI. }\label{mablation}
\begin{tabular}{lcc}
\toprule
Scale & MAE & MSE \\
\midrule
5, 7, 11  & 4.8 & 6.2 \\
3, 7, 11  & 4.7 & 6.2 \\
3, 5, 7  & 4.6 & 6.1 \\
\midrule
\textbf{3, 5, 7, 9}  & \textbf{4.4} & \textbf{5.9} \\
\bottomrule
\end{tabular}
\end{minipage}
\end{minipage}
\subsection{Comparison with State-of-the-Art Approaches}
Our approach demonstrates state-of-the-art performance across three key benchmarks, shown in Tab.\ref{counting}:
On ADI, we achieves new records with MAE=4.1 and MSE=5.7, surpassing previous best by 51.2\%/51.3\%. For MBM, our approach establish new benchmarks at MAE=0.1 / MSE=0.3, improving upon existing best by 92.9\%/85.7\%. On VGG, our model reduces errors to MAE=2.0 / MSE=2.7, cutting prior records by 51.2\%/54.2\%.

While our approach achieves competitive MAE=1.7 / MSE=3.0 on DCC (surpassing $\sim$90\% baselines), its slight gap compared to DCL stems from dataset-specific factors of DCC. The performance gap primarily stems from three intrinsic DCC characteristics: 1) Extreme scale variance introduces multi-resolution conflicts with our fixed-scale feature extraction. 2) Distinct cell lineages (embryonic/monocyte/cancer) require specialized morphological priors that our cross-dataset unified encoder underserves. 3) Sparse cell distributions contradict our framework's strength in modeling dense cell overlaps through positional context reasoning. DCC's average 34 cells/image benefits methods with explicit density estimation modules. Our position-based counting framework shows relative advantage in dense scenarios (ADI/MBM/VGG) over sparse distributions.
\begin{figure}[!t]
    \centering
    \includegraphics[width=\textwidth]{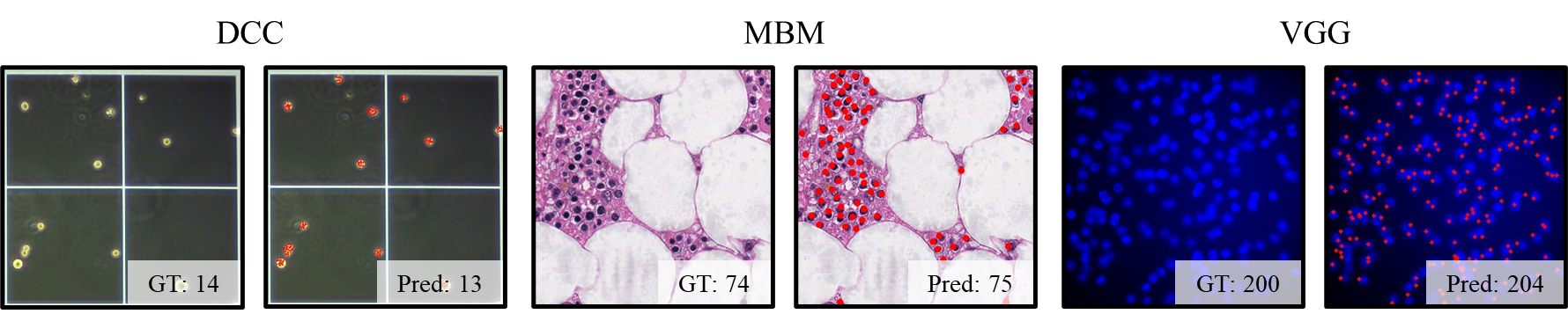}
    \caption{Visualization across multiple datasets.}
    \label{visual}
\end{figure}

\subsection{Ablation Study}
Our ablation study on the ADI dataset conclusively demonstrates the critical contributions of the proposed KHM Algorithm and MDGC modules, shown in Tab.\ref{ablation}, Tab.\ref{kablation} and Tab.\ref{mablation}. When implementing KHM - designed to adaptively match cellular neighborhoods in dense adipocyte clusters - we observe dramatic MAE/MSE reductions from 9.0/13.1 (Baseline) to 4.6/6.1, representing 48.9\% and 53.4\% error decreases. The integration of MDGC, which handles size-varying cells through deformable receptive fields, improves results to 4.4/5.9 (51.1\%/55.0\% improvements). The complete model achieves optimal performance at 4.1 MAE and 5.7 MSE, delivering 54.4\%/56.5\% overall error reduction. In the ablation studies of KHM and MDGC, after multiple parameter modifications, we found that the modules could achieve the best performance when k = 5 and scale = \{3, 5, 7, 9\}.

%This breakthrough directly addresses ADI's unique challenges: 1) Ultra-dense adipocyte packing (90\%+ spatial occupancy) requiring precise neighborhood modeling (KHM's role); 2) Significant cell size polymorphism demanding scale-adaptive feature extraction (MDGC's function); 3) Low intercellular contrast from histological staining artifacts mitigated by Gaussian-based confidence weighting.
\section{Conclusion}
In this paper, we propose three innovative technologies for dense cell counting: KHM, MDGC, and GFD. These breakthroughs systematically address critical challenges in cell image analysis, including matching inaccuracies, limited feature extraction, and spatial information loss.

KHM dynamically adjusts matching ranges using Gaussian weighting, enabling intelligent label assignment in overlapping regions. MDGC enhances cell recognition through deformable kernels and multi-scale fusion, effectively handling diverse cell morphologies. GFD preserves precise counting via selective feature enhancement and geometric constraints. Together, these innovations overcome traditional limitations in density sensitivity, shape adaptability, and spatial degradation.

The framework holds significant value for biomedical applications, providing reliable automated analysis for pathological diagnosis. Experiments demonstrate its state-of-the-art performance across multiple public datasets, particularly excelling in complex scenarios with extreme cell density.

\newpage

\begin{comment}  %% removed for anonymized MICCAI 2025 submission.
    
    % The following acknowledgement and disclaimer sections should be removed for the double-blind review process.  
    % If and when your paper is accepted, reinsert the acknowledgement and the disclaimer clause in your final camera-ready version.

\begin{credits}
\subsubsection{\ackname} A bold run-in heading in small font size at the end of the paper is
used for general acknowledgments, for example: This study was funded
by X (grant number Y).

\subsubsection{\discintname}
It is now necessary to declare any competing interests or to specifically
state that the authors have no competing interests. Please place the
statement with a bold run-in heading in small font size beneath the
(optional) acknowledgments\footnote{If EquinOCS, our proceedings submission
system, is used, then the disclaimer can be provided directly in the system.},
for example: The authors have no competing interests to declare that are
relevant to the content of this article. Or: Author A has received research
grants from Company W. Author B has received a speaker honorarium from
Company X and owns stock in Company Y. Author C is a member of committee Z.
\end{credits}

\end{comment}

\end{document}